\documentclass{article}

\usepackage[utf8]{inputenc}
\usepackage{caption}
\usepackage{graphicx}
\usepackage{amsmath}
\usepackage{amsthm}
\usepackage{booktabs}
\usepackage{algorithm}
\usepackage{algorithmic}
\usepackage{amsfonts}
\usepackage{authblk}
\usepackage[a4paper, total={6.2in, 8.5in}]{geometry}
\usepackage[T1]{fontenc}
\usepackage{tgtermes}

\usepackage{url}
\usepackage{biblatex}
\title{Feature-based Graph Attention Networks Improve Online Continual Learning}

\author[1,2]{Adjovi Sim}
\author[2]{Zhengkui Wang}
\author[1,2]{Aik Beng Ng}
\author[1]{Shalini De Mello}
\author[1]{Simon See}
\author[1]{Wonmin Byeon}
\affil[1]{NVIDIA}
\affil[2]{Singapore Institute of Technology}

\date{}

\begin{document}

\maketitle

\begin{abstract}
\noindent Online continual learning for image classification is crucial for models to adapt to new data while retaining knowledge of previously learned tasks. This capability is essential to address real-world challenges involving dynamic environments and evolving data distributions. Traditional approaches predominantly employ Convolutional Neural Networks, which are limited to processing images as grids and primarily capture local patterns rather than relational information. Although the emergence of transformer architectures has improved the ability to capture relationships, these models often require significantly larger resources. In this paper, we present a novel online continual learning framework based on Graph Attention Networks (GATs), which effectively capture contextual relationships and dynamically update the task-specific representation via learned attention weights. Our approach utilizes a pre-trained feature extractor to convert images into graphs using hierarchical feature maps, representing information at varying levels of granularity. These graphs are then processed by a GAT and incorporate an enhanced global pooling strategy to improve classification performance for continual learning. In addition, we propose the rehearsal memory duplication technique that improves the representation of the previous tasks while maintaining the memory budget. Comprehensive evaluations on benchmark datasets, including SVHN, CIFAR10, CIFAR100, and MiniImageNet, demonstrate the superiority of our method compared to the state-of-the-art methods.
\end{abstract}

\section{Introduction}
Online continual learning (OCL) is a vital paradigm in machine learning, enabling models to incrementally adapt to new data while preserving knowledge from previously learned tasks \cite{wang2024comprehensive,huo2024non}. This approach is particularly important in dynamic environments with evolving data distributions, as it addresses the challenge of catastrophic forgetting — where models lose prior knowledge when updated with new information \cite{McCloskey1989CatastrophicII_castastrophicforgetting,vandeVen2019ThreeSF_threescenarios}. OCL is especially critical for image classification in numerous real-world applications scenarios \cite{vandeVen2019ThreeSF_threescenarios}, such as autonomous vehicles needing to adapt to new environments, facial recognition needing to recognize new images over time, and other scenarios such as video surveillance, and climate monitoring, where the ability to learn and adapt continuously is essential to maintaining performance and relevance.

Previous approaches to mitigating catastrophic forgetting have heavily relied on Convolutional Neural Networks (CNNs) \cite{jung2023new_mufan,vandeVen2019ThreeSF_threescenarios}. CNNs hierarchically aggregate local features to generate final image representations. 
However, they fail to capture the relationships between the parts of the image. Furthermore, they lack the capability of the representation update, as features are simply accumulated toward the final representation rather than being dynamically refined based on surrounding regions. Although transformer architectures offer a potential solution through their attention mechanisms, which can model the relationship between features, their substantial computational requirements are large.

In contrast, Graph Neural Networks (GNNs) enable dynamic representation updates through message passing between neighboring nodes \cite{zhou2021overcoming}. It captures relational information across different image segments, with an effective receptive field to n-hops, where n corresponds to the network depth. Through iterative aggregation and update operations, GNNs can learn diverse graph structures specific to different tasks and can inherently incorporate global structural information.

However, GNNs on images require a graph construction phase, as images are inherently non-graph structured data. Various graph construction methods exist beyond the naive pixel-to-node mapping, such as Simple Linear Iterative Clustering (SLIC) \cite{fey2019fastgraphrepresentationlearning_pyg}. Although SLIC is frequently employed in image-based GNN benchmarking studies, each node's feature representation is maximally five values, RGB color values and XY spatial coordinates, which restricts the final graph representation.

A fundamental challenge in continual learning is mitigating catastrophic forgetting — the severe performance deterioration on previously acquired tasks. Existing continual learning approaches focus on optimizing the stability-plasticity trade-off, aiming to preserve performance on previous tasks while maintaining the adaptability to new tasks \cite{ABRAHAM200573_stabilityplasticity}. This optimization becomes particularly critical in online learning scenarios, where data samples can only be processed once.

While DGN \cite{Carta2021CatastrophicFI_dgn} introduces image-based GNN for continual learning to solve the catastrophic forgetting issue, the stability-plasticity dilemma persists. Therefore, we propose attention-based GNN architecture for online continual learning on complex multi-task scenarios.  This architectural enhancement enables the model to capture dynamic relationships between image segments and learn adaptive task-specific graph structures.

In this paper, we propose Feature-based Graph Attention Networks (FGAT), a novel architecture that transforms images into graph representations in a high-dimensional feature space. We utilize a pre-trained feature extractor, extracting features at multiple levels of granularity, so each node encodes multi-scale feature information, modeling complex relationships between features. The extracted features are processed using a Graph Attention Network (GAT), which enables dynamic node representation updates based on local neighborhood context. We also introduce a custom weighted global mean pooling mechanism that aggregates information from all nodes based on their relative importance. 
Finally, to enhance stability (maintaining performance on the past tasks), we incorporate a rehearsal buffer that maintains and utilizes a subset of previous task data. We further introduce a novel rehearsal memory duplication strategy that maximizes the utility of the rehearsal memory while constraining the memory budget, emulating joint-task training without having access to the entire set of past task images. It effectively enhances the representational quality of previous tasks while training new tasks, thereby maintaining robust performance across tasks.

Our key contributions are as follows:
\begin{enumerate}
    \item We propose Feature-based Graph Attention Networks (FGAT), a novel architecture for online continual learning that leverages multi-scale feature representations and the attention mechanism. By utilizing a pre-trained feature extractor, FGAT enables each node to encode rich, hierarchical feature information, effectively modeling semantic dependencies. 
    \item We develop an attention-based graph structure that allows the model to capture complex topological patterns and task-specific graph structures for online continual learning. Additionally, we introduce a learnable weighted global mean pooling mechanism that optimizes node contributions to the final representation based on their relative importance. 
    \item We propose a novel rehearsal memory duplication strategy that addresses the fundamental challenge of memory imbalance in rehearsal-based approaches. It enhances the representation quality of past tasks while maintaining the new task performance without increasing the memory budget.  
    \item We evaluate FGAT on four benchmark datasets (SVHN, CIFAR10, CIFAR100, and MiniImageNet) and compare it with existing CNN and GNN-based continual learning methods. The results demonstrate that our proposed method outperforms current state-of-the-art approaches.   
\end{enumerate}  

\section{Related Work}
\subsection{Continual Learning}
Continual Learning (CL) is a paradigm designed to enable models to sequentially learn a series of tasks while preserving performance on previously learned tasks. This capability addresses the phenomenon of catastrophic forgetting, a common issue in standard training approaches where new learning overwrites prior knowledge. CL is typically categorized into three main settings: Task-Incremental, Domain-Incremental, and Class-Incremental \cite{vandeVen2019ThreeSF_threescenarios}. Among these, Class-Incremental Learning is particularly challenging, as it requires the model to solve tasks without explicit task identifiers and to classify across all encountered classes, thereby minimizing reliance on task-specific information.

In CL, pre-trained models are often employed as feature extractors, leveraging representations generated by the final feature map. These approaches can be broadly divided into two categories: those that treat the pre-trained model as a trainable backbone and those that use it as a fixed feature extractor \cite{jung2023new_mufan}. However, most methods in the latter category primarily focus on the final layer representation, neglecting the potential benefits of multi-granularity feature representations.

Recent work, such as MuFAN, tackles this limitation by leveraging feature maps from multiple layers to create richer representations for online continual learning \cite{jung2023new_mufan}. This approach primarily utilizes convolutional layers to produce multi-scale representations. In contrast, our work employs GNNs to advance this concept further. By integrating multi-granularity features into a graph construction process, we create information-rich graph representations of images, facilitating more effective learning in OCL scenarios.

\subsection{Graph Based Continual Learning}
Graph Neural Networks (GNNs) have garnered significant attention in the field of continual learning, demonstrating considerable promise. For instance, TWP \cite{Liu2020OvercomingCF_twp} utilizes topological information to stabilize the training process by measuring the attention coefficients in a Graph Attention Network (GAT). However, existing studies predominantly focus on graph-structured data, and can not be directly used for CL for image classification.

Recent research works have explored the use of GNNs for continual learning in image classification. DGN \cite{Carta2021CatastrophicFI_dgn} employs a static clustering method and evaluates its performance on MNIST and CIFAR10 datasets. Similarly, GCL \cite{tang2021graphbased_gcl} models image batches using random graphs, departing from conventional image-based methods. However, to the best of our knowledge, this is the first work to explore graph attention networks for continual learning in image classification.

\section{Background}
\subsection{Image to Graph}
Images are inherently structured as grid-based data rather than graph-based. Consequently, transforming an image from its grid representation to a graph representation is necessary. A straightforward approach involves representing each pixel as a node and establishing edges between adjacent pixels. However, this method is highly inefficient, as neighboring pixels with similar attributes, such as color, are redundantly represented as separate nodes, leading to unnecessary complexity.

One approach to tackling this issue is with SuperPixel methods, such as Simple Linear Iterative Clustering (SLIC) \cite{article_slic} which aggregates neighborhood information into each node instead, having a node represent a spatial area. This is done by seeding initial cluster centers, before iteratively moving the cluster centers to the position of lowest color gradient in CIELAB space, associating each neighborhood with a cluster, and repeating until the distance moved by the cluster centers drops below a specified threshold \cite{article_slic}. However, while this tackles the issue of repeated information, the representation is kept to color and spatial information, R/G/B/X/Y. In contrast, more information-rich representations are possible, through the use of a pre-trained feature extractor to represent the image in feature space instead. By constructing the image in feature space instead, we are able to capture information across hundreds of feature maps, yielding a more information-rich representation as compared to color. MuFAN \cite{jung2023new_mufan} leverages this to enable faster classifier training, taking advantage of multi-scale feature maps for downstream CNN based classifier training.

\subsection{Graph Neural Networks}
A primary attribute of GNNs is the layers used within, as they affect the main operation of GNNs, which is the message passing and update step. One such family of layers are graph convolutional layers. In the simplest form, this would be $Node' = Aggregate(Neighbour(Node)) \times Node$ which simply aggregates the information from neighboring node into the node to be updated. 

Trainable types of layers generally evolve from this main formula, such as GraphConv \cite{Morris2018WeisfeilerAL_graphconv}, which adds a set of trainable weights for both the aggregate as well as node to be updated. Another component which can be swapped out would be the aggregation operation. Switching out the summation operation in GraphConv \cite{Morris2018WeisfeilerAL_graphconv} to a mean operation would yield the general update formula used for SageConv \cite{Hamilton2017InductiveRL_graphsage}.
 
Instead of changing the aggregation operation, the weighing of neighboring nodes can be adjusted as well. GCNConv \cite{kipf2017semisupervised_gcnconv} iterates on the general formulation by including the degree of the nodes. GATConv \cite{veličković2018graph_gat} improves on the neighbor nodes weightage by adding a learnable attention mechanisms between the neighboring nodes and node to be updated, as well as self-attention for the node to be updated. GATv2Conv \cite{brody2022how_gatv2} further improves upon GATConv by re-arranging the order of operations to allow the attention mechanism to perform better based on the inputs.

\subsection{Graph Pooling}
\begin{figure}
    \centering
    \includegraphics[width=\linewidth]{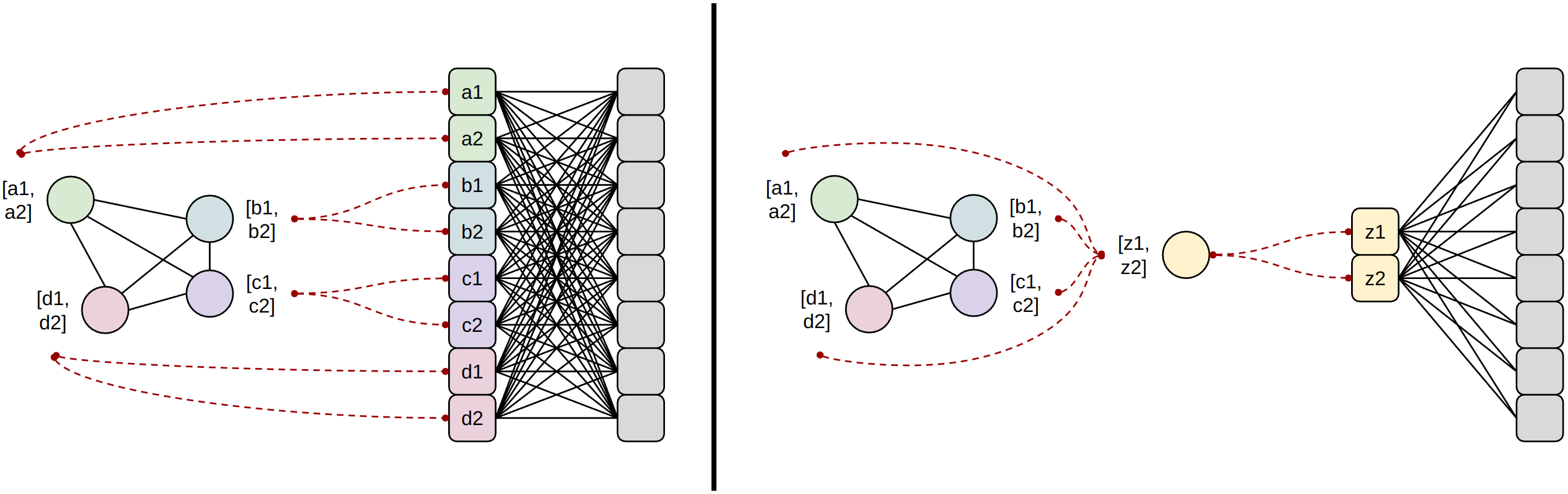}
    \caption{Utilization of entire graph for classification (left), utilization of global pooling before classification (right)}
    \label{fig:graph_pooling}
\end{figure}

In this step, the graph representation is flattened and transformed into a vector for input into the classification block. This can be done in a straightforward manner as illustrated in Figure \ref{fig:graph_pooling} (left), whereby the entire graph is used as input into the classification block. However, this results in a large input, and importantly, requires the number of nodes be fixed as it directly affects the input size of the classification block. 

The number of nodes can instead be reduced through global pooling operations, aggregating the entire graph's information into a single node, illustrated in Figure \ref{fig:graph_pooling} (right).  
DGN \cite{Carta2021CatastrophicFI_dgn} does global pooling by using max operations, with other options being add and mean operations \cite{fey2019fastgraphrepresentationlearning_pyg}.
While utilizing a global mean pooling operation allows for all nodes to contribute to the final representation for classification, it treats all nodes with equal importance. Given a repeated background pattern, this can lead to an overly high impact on the final representation. Therefore, we propose weighted global mean pool. This allows each node to contribute to the final representation, but at varying levels based on the other nodes in the graph. With different tasks and therefore different distribution of node representations, this allows for emphasis of task important nodes in the final representation.
This final representation can then be passed into a classifier block for classification.

\subsection{Distillation Techniques}
Beyond just architecture design, additional continual learning specific techniques can be utilized to help mitigate catastrophic forgetting. A family of such techniques, called distillation techniques, aims to utilize a saved copy of the model prior to starting a new task, to distill information from prior tasks to the new task. One such technique is Learning without Forgetting (LwF) \cite{Li2016LearningWF_lwf} which utilizes the saved copy to obtain soft targets using current task images. This allows for regularization of the model in that manner to not stray from the learnt output distribution using primarily current task images.

\section{Method}

\begin{figure}
    \centering
    \includegraphics[width=\linewidth]{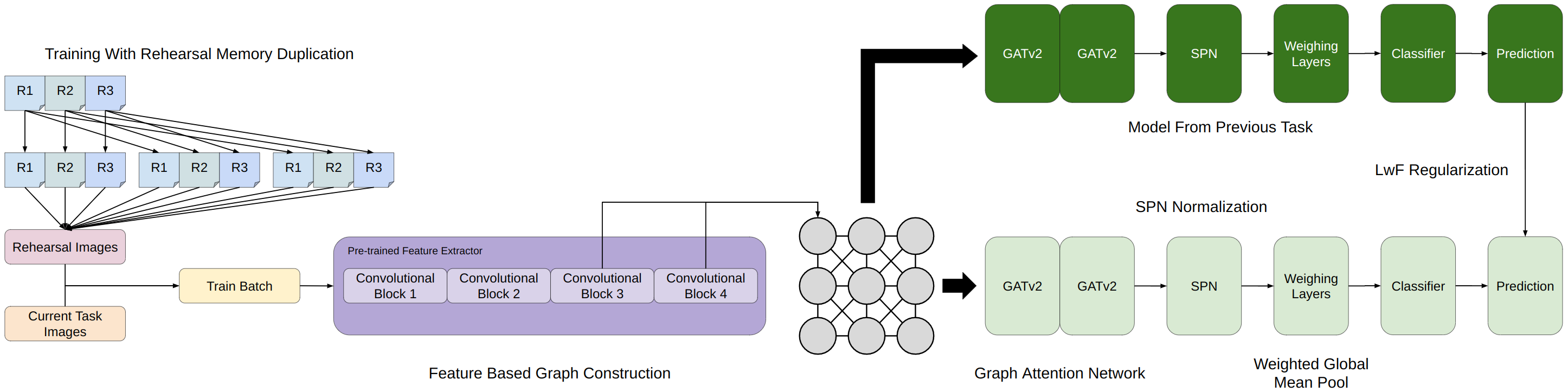}
    \caption{Overview of FGAT, illustrated with three total rehearsal images - [R1, R2, R3]}
    \label{fig:method_illustration}
\end{figure}

Figure \ref{fig:method_illustration} provides an overview of the proposed FGAT framework. FGAT begins by leveraging a feature extractor, such as a pre-trained CNN, to construct feature-based graphs from input images (described in Section \ref{subsec:feature_based_graph_construction}). This is followed by the application of a novel graph attention network (detailed in Section \ref{subsec:graph_attention_network}) and a weighted global mean pooling mechanism for continual learning (outlined in Section \ref{subsec:w_gmp}). Furthermore, FGAT incorporates SPN normalization for the GNN (explained in Section \ref{subsec:normalization}) and utilizes rehearsal memory duplication techniques to optimize memory usage while increasing representation of previous tasks (described in Section \ref{subsec:rehearsal_duplication}).

\subsection{Feature Based Graph Construction} \label{subsec:feature_based_graph_construction}
The first step in our proposed approach involves transforming an image into a graph representation. A straightforward method could operate directly on raw image data, which inherently comprises five attributes: three color channels (RGB) and two spatial coordinates (XY). However, by utilizing a Convolutional Neural Network (CNN) based feature extractor, such as ResNet, we can extract feature maps from the image, enabling graph construction in a richer feature space.

Using a pre-trained CNN as the feature extractor, we construct the graph in a high-dimensional feature space that captures semantic and contextual information. For instance, when ResNet18 is employed as the backbone architecture, its final convolutional block outputs a 512-dimensional feature vector. Each pixel location in the feature map is treated as a node, with features representing the semantic information aggregated across all feature maps. This approach encodes complex patterns, textures, and hierarchical representations essential for constructing a robust graph.

To enhance the representation, features from multiple layers of the CNN are utilized, incorporating both fine-grained and high-level semantic information. This multi-level feature aggregation approach promotes increased plasticity in online learning scenarios by integrating hierarchical representations \cite{jung2023new_mufan}. Specifically, we employ feature maps from the final two convolutional blocks of the CNN feature extractor in our implementation.

When reconciling feature maps of different resolutions, we avoid using linear or bilinear upsampling techniques, as these assume a smooth interpolation of feature values, which may not align with semantic relationships. For instance, a feature map representing a detected circle in one region of the image and no circle in another region does not imply a gradual progression of ``less clear" circles between these areas. Instead, each pixel independently contributes to the feature representation. Therefore, we upsample feature maps by repeating each feature value by a factor corresponding to the resolution difference. This method preserves the semantic integrity of the features and avoids introducing artificial interpolations. As a concrete case, upsampling [1, 4] would result in [1, 1, 4, 4].

The feature assignment is performed by mapping each node to its corresponding spatial location in the hierarchical feature maps. Specifically, for each node, we concatenate the feature values from all corresponding positions across the multi-scale feature maps, together with the spatial coordinates (X, Y). This results in a comprehensive node representation that encodes both multi-granular semantic features and explicit spatial information. The edges are then constructed between nodes based on their spatial proximity through a k-nearest neighbor (k-NN) algorithm.

\subsection{Graph Attention Network} \label{subsec:graph_attention_network}
Consider that in CL for image classification, the tasks can differ significantly, and indiscriminate addition of knowledge can lead to catastrophic forgetting. We argue that adding the attention mechanism to the GNN is essential to mitigate forgetting. Attention allows the GNN to focus on the most relevant neighbouring nodes during the node update step for a given task, reducing interference from less relevant information. Meanwhile, by assigning attention weights, the GNN can dynamically adapt to the varying importance of features across different tasks, and can also prioritize preserving critical features and relations learned from previous tasks. 

To do so, in this paper, we propose to use GATv2 \cite{brody2022how_gatv2} for learning the feature graphs in FGAT. 
While Graph Attention Networks (GAT) \cite{veličković2018graph_gat} enhances the message passing mechanism in Graph Neural Networks by introducing learnable attention weights for neighbor aggregation, GATv2 \cite{brody2022how_gatv2} further refines this approach through a reorganization of computational operations, enabling attention scores that are conditioned on both the query and key. GATv2 computes the normalized importance between nodes i and j as follows:
\begin{align}
    \alpha_{i,j} = \text{softmax}_j (a^\top \text{LeakyReLU}(w \cdot [h_i | h_j]))
\end{align}
where $h_i$ and $h_j$ are the node representations for node i and j respectively. $a$ and $w$ are learnable weights. The node representation is then updated:
\begin{align}
    h_i = \sigma (\sum_{j \in N_i}\alpha_{ij} \cdot wh_j)
\end{align}
whereby $\sigma$ represents a non-linearity operation.
This dynamic attention mechanism allows nodes to selectively aggregate information from their neighbors based on learned importance weights.

The attention-driven graph structure is pivotal in online continual learning for image classification due to its adaptive feature learning capabilities. The dynamic node representation allows nodes to update their representations using learned attention weights selectively aggregating neighborhood features based on task relevance. This ensures that only the most pertinent information influences the final representation. The dynamic attention mechanism adapts information flow between nodes to meet each task's requirement. This prevents catastrophic forgetting of prior tasks while incorporating task-relevant neighborhood information into updates.

\subsection{Weighted Global Mean Pool} \label{subsec:w_gmp}
\begin{figure}
    \centering
    \includegraphics[width=\linewidth]{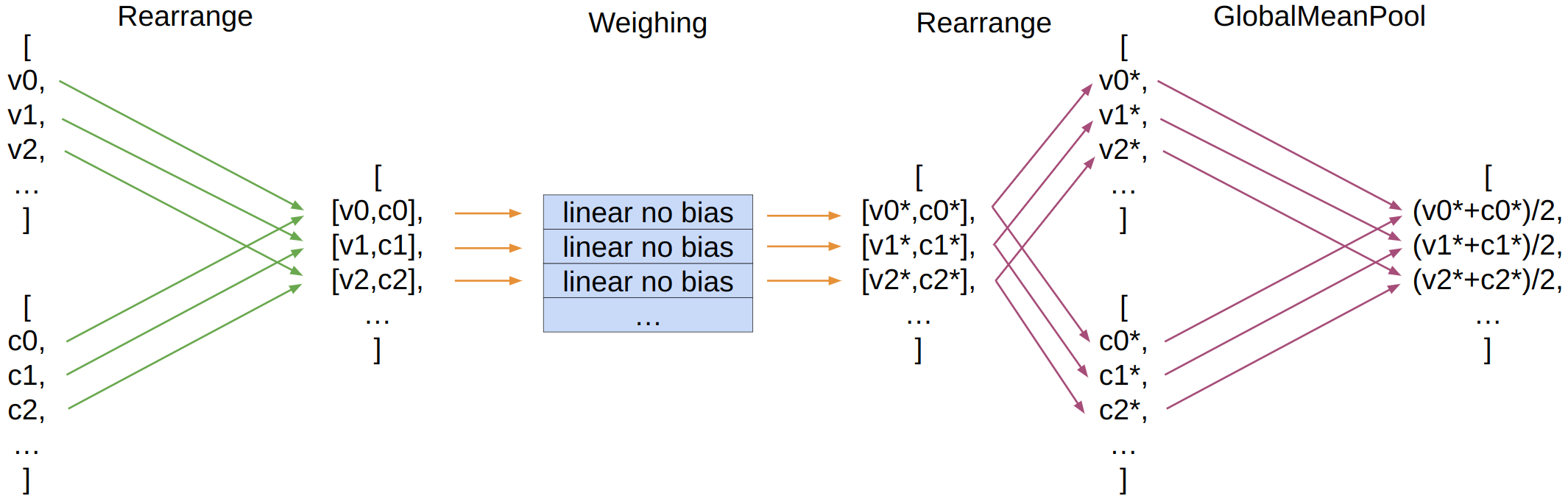}
    \caption{Process of performing weighted global mean pool}
    \label{fig:weighted_global_mean_pool}
\end{figure}

Instead of standard global pooling, we employ a channel-wise weighted pooling strategy. An illustration of this approach is shown in Figure \ref{fig:weighted_global_mean_pool}. We first transform the graph representation such that the channel values across nodes are in the same dimension. We then independently weigh each set of channel values using a linear layer, before reconstructing the original graph representation format to obtain a weighted representation. This is then passed into a mean pooling operation. This approach preserves the relative importance of different nodes in the final representation. Also, it adaptively determines the contribution of each node before the mean pooling operation.

\begin{figure}
    \centering
    \includegraphics[width=0.5\linewidth]{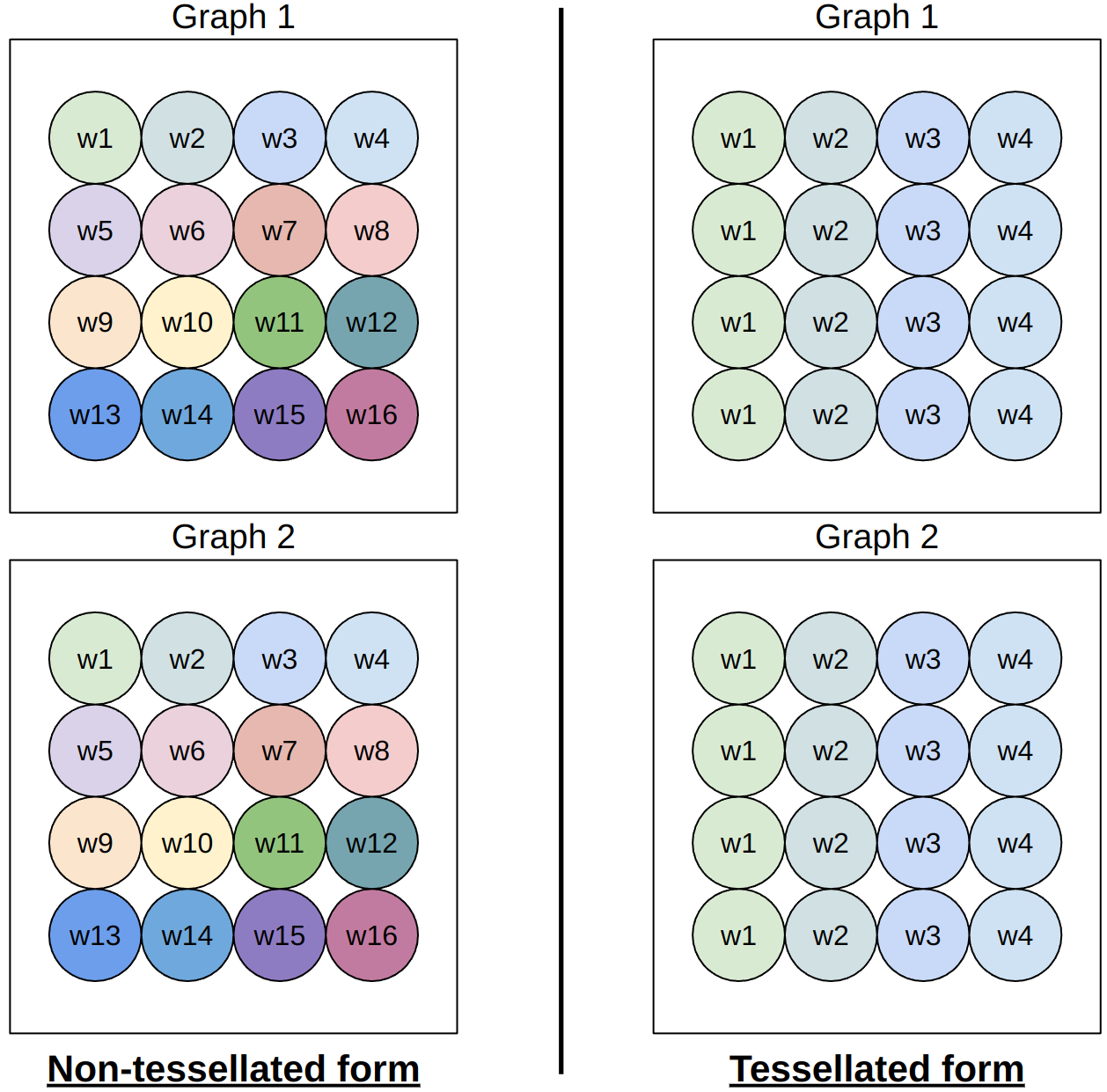}
    \caption{In the non-tessellated form, each node in the graph receives a separate weight when passed through the weighing layer. In the tessellated form, sub-sets of the graph, in this case sets of 4 nodes receives separate weights, which are shared across the rest of the graph.}
    \label{fig:tessellated_form}
\end{figure}

While this allows for the entire set of nodes to be weighed against each other, it necessitates that the size of the graph be fixed to determine the number of weights needed. We utilize a version whereby instead of weighing all nodes at the same time, we operate on smaller sub-sets (specifically a quarter) at a time using the same smaller weighing layer, allowing for possible graph size growth in sets of the patch size. We name this form the tessellated form, while the entire graph form the non-tessellated form, and illustrate how the division of a graph into sub-sets as well as the shared weights in Figure \ref{fig:tessellated_form}. In Section \ref{subsec:ablation}, we show the comparison between the standard and the tessellated form of weighted global pooling.

\subsection{Node Normalization} \label{subsec:normalization}
We adopt the SPN normalization block introduced in \cite{jung2023new_mufan} for Graph Neural Network (GNN).
Instead of feature map outputs from Convolutional Neural Networks (CNNs), we treat the entire graph as the feature map to be normalized. In doing so, we remap nodes to be akin to pixel locations, and node channels to be the pixel value across feature nodes. We utilize one such block post GNN and before performing weighted global mean pooling.

\subsection{Training With Rehearsal Memory Duplication} \label{subsec:rehearsal_duplication}
In rehearsal-based continual learning, a subset of data from previous tasks is maintained in memory for replay during future learning. However, as the size of this subset of data is less than that of the current task, there is an inherent representation imbalance between current and previous tasks. While existing methods such as ER \cite{Chaudhry2019ContinualLW_er} address this by sampling from the replay buffer with replacement during training, we propose a simple yet effective solution: \textit{duplicating rehearsal samples and incorporating them into the training set to emulate joint-task training}. The memory buffer, as in other rehearsal-based methods, stores a limited number of image samples from previous tasks. During training, each stored sample is duplicated multiple times. This strategy significantly enhances the representation of previous tasks without increasing the memory budget, while requiring fewer total images per past task compared to the current task. 

Given $T$ total tasks, the image classes $C$ can be divided into $T$ tasks, represented as $T = \{t_1, t_2, ...\} = \{\{c_1, \dots\}, \{c_\alpha, ...\}, ...\}$, where each image belonging to the task is denoted as $i_\beta \in I_t, i_\beta = \{x_\beta, y_\beta\}$ such that $y_\beta = c_\beta \in t$. The graph construction step is represented as $f$ and graph neural network and classifier block trained till task $t$ is represented as $g_t$ and $p_t$ respectively, while the rehearsal memory accumulated up to task $t$ is denoted as $R_t$.
The classification cross-entropy loss is defined as follows: 
\begin{align}
    L_{CE} = \sum_{i=0} l_{ce}(p_{t}(g_{t}(f(x_i))),y_i) \in I_t \cup R_t
\end{align}
where $l_{ce}$ represents cross-entropy. In addition to utilizing rehearsal memory, we utilize Learning without Forgetting (LwF) on the combined train set \cite{Li2016LearningWF_lwf} as well, passing each training batch into the saved copy from the previous task and utilizing the output as a soft target for the current batch. This distillation loss is given as:
\begin{align}
    L_{DL} = \sum_{i=0} l_{kl}(p_i^\text{old},\text{log}(p_i^\text{new}))
\end{align}
where $p_i^{old}$ and $p_i^{new}$ are the outputs of the models from the previous and new tasks for the input image $i$: $p_i=\text{softmax}(p_{t}(g_{t}(f(x_i)))/T)$ and $l_{kl}$ is the Kullback-Leibler divergence. T is the temperature hyper-parameter. 
The overall loss function for training is defined as:
\begin{align}
    L = L_{CE} + \alpha_{DL} L_{DL}
\end{align}
where $\alpha_{DL}$ is a hyper-parameter controlling knowledge distillation strength.

\section{Experiments}
We evaluate FGAT in an online task-incremental setting, where no task identity is provided, and the model is required to classify across all image classes. FGAT's performance is compared against CNNs utilizing graph representations, CNNs with pre-trained feature extractors, and GNNs. The experiments are evaluated on 5-Task Street View House Numbers (SVHN) \cite{SVHN,tang2021graphbased_gcl}, 5-Task CIFAR10 \cite{article_cifar,tang2021graphbased_gcl}, 20-Task CIFAR100 \cite{article_cifar,LopezPaz2017GradientEM_gem_splitcifar100}, and 20-Task MiniImageNet \cite{ravi2017optimization_miniimagenet,Vinyals2016MatchingNF_miniimagenet,jung2023new_mufan}. 

\subsection{Dataset}
This section provides an overview of the four benchmark datasets utilized in the experiments.

\textbf{Street View House Numbers (SVHN).} 
Street View House Numbers (SVHN) \cite{SVHN} consists of 10 images classes, 0 to 9, with Split-SVHN splitting the image classes into 5 tasks \cite{tang2021graphbased_gcl}. The images are originally of size 32x32 and are class-imbalanced. Critically, for this dataset, the method of acquisition results in a particular characteristic. We observe that as the bounding boxes are extended to form squares windows, with the goal of avoiding aspect ratio issues \cite{SVHN}, some images  include artifacts of neighboring numbers adjacent to the main central number.

\textbf{CIFAR10.}
CIFAR10 \cite{article_cifar} consists of 10 image classes as well, with Split CIFAR10 also splitting the image classes into 5 tasks \cite{tang2021graphbased_gcl}. The images are originally of size 32x32 and are class-balanced.

\textbf{CIFAR100.}
CIFAR100 \cite{article_cifar} consists of 100 image classes, with Split CIFAR100 splitting the image classes into 20 tasks \cite{LopezPaz2017GradientEM_gem_splitcifar100}. The images are originally of size 32x32 and are class-balanced.

\textbf{MiniImageNet.}
MiniImageNet \cite{ravi2017optimization_miniimagenet,Vinyals2016MatchingNF_miniimagenet} consists of 100 image classes as well, with Split MiniImageNet also splitting the image classes into 20 tasks \cite{jung2023new_mufan}. As the images are originally from the ImageNet dataset, the images are originally of a higher resolution of 224x224. However, we adopt the MiniImageNet downscale to 84x84 before re-upscaling it to 224x224 for processing to maintain the intended level of image fidelity as opposed to directly working with the 224x224 size images. The images are class-balanced.

\subsection{SuperPixel Construction} 
\begin{figure}
    \centering
    \includegraphics[width=\linewidth]{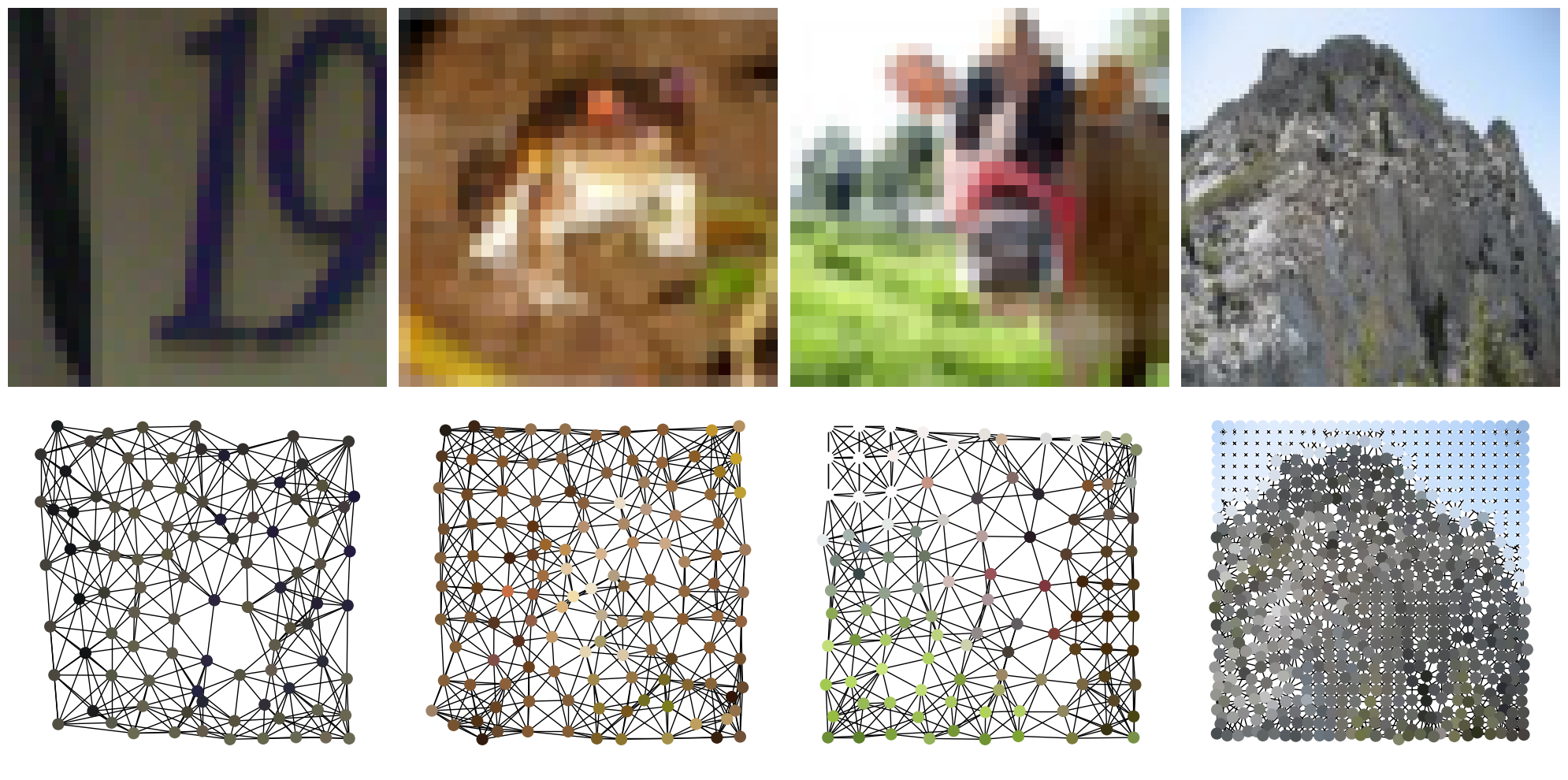}
    \caption{Sample images from, in order, SVHN, CIFAR10, CIFAR100, MiniImageNet (top) and the corresponding SuperPixel graph (bottom)}
    \label{fig:slic_visualization}
\end{figure}

In cases where superpixel representations generated using SLIC are unavailable in PyTorch Geometric \cite{fey2019fastgraphrepresentationlearning_pyg}, the ToSLIC method within PyTorch Geometric was employed, with the maximum number of nodes configured as detailed in Table \ref{tab:slic_max_nodes}. For datasets with image sizes comparable to those with existing representations, the same maximum node count was applied. For MiniImageNet, the maximum number of nodes was estimated linearly by extrapolating from the values used for MNIST and CIFAR10. Specifically, we utilized $75 + ((100 - 75)) \times (84^2 - 32^2) / (32^2 - 28^2))) = 703.33 \approx 700$. An illustration of sample images and corresponding graphs is shown in Figure \ref{fig:slic_visualization}.

\begin{table}[t]
    \centering
    \begin{tabular}{ccc}
        \toprule
        Dataset & Image Resolution & Max Nodes \\
        \midrule
        MNIST & 28x28 & 75 \\
        SVHN & 32x32 & 100 \\
        CIFAR10 & 32x32 & 100 \\
        CIFAR100 & 32x32 & 100 \\
        MiniImageNet & 84x84 & 700 \\
        \bottomrule
    \end{tabular}
    \caption{Simple Iterative Linear Clustering (SLIC) Max Nodes}
    \label{tab:slic_max_nodes}
\end{table}

\subsection{Architecture}
FGAT utilizes features from the last two convolution blocks of a frozen pre-trained ResNet18 network, with the smaller feature maps being stretched to the size of the larger feature maps, resulting in repeated values for the elements in the smaller feature map, $i \rightarrow{} e_1 \in \mathbb{R}^{C_1 \times W_1 \times H_1}, e_2 \in \mathbb{R}^{C_2 \times W_2 \times H_2} \rightarrow e_1 \in \mathbb{R}^{C_1 \times W_1 \times H_1}, e_2 \in \mathbb{R}^{C_2 \times W_1 \times H_1} \rightarrow e_{final} \in \mathbb{R}^{(C_1 + C_2) \times W_1 \times H_1}$ whereby C is the number of channels of the feature map, and W and H are the width and height respectively. To construct the edges, we utilize a spatial k-NN approach similar to SLIC, connecting each node to its nearest 8 spatial neighbors.

For the feature extractor, the implementation provided by PyTorch \cite{Ansel_PyTorch_2_Faster_2024_pytorch} along with the provided trained weights from training on ImageNet were used. The model's output from the last 2 convolutional blocks were used, yielding an output of $256 \times 14 \times 14$ and $512 \times 7 \times 7$ before re-scaling and stacking. 2 GATv2Conv from \cite{fey2019fastgraphrepresentationlearning_pyg} was used for the Graph Neural Network (GNN), with 4 attention heads used for SVHN and CIFAR10, and 3 attention heads used for CIFAR100 and MiniImageNet, and a base number of channels of 128. One SPN normalization block, with implementation referenced from the code base made available by \cite{jung2023new_mufan}, though adapted to work with GNNs as detailed earlier, is used after the GNN layers. The classifier is comprised of 2 linear layers, with 128 features.

\subsection{Hyper-parameters}
Learning without Forgetting (LwF) \cite{Li2016LearningWF_lwf} with reference implementation from \cite{Carta2021CatastrophicFI_dgn} was used. LwF hyper-parameter search was done using a random search approach with alpha values ranging from 0.25 to 2.50 and temp values ranging from 0.25 to 2.50. Duplication level was set at 20 for SVHN and CIFAR100, and at 15 for CIFAR10, and MiniImageNet. Models were trained using the Adam optimizer, with a learning rate of 0.001. Following \cite{tang2021graphbased_gcl}, we utilized 5 rehearsal/class, 10 rehearsal/task for SVHN and CIFAR10. We also follow \cite{jung2023new_mufan} and utilize 10 rehearsal/class, 50 rehearsal/task for CIFAR100 and MiniImageNet. 

\subsection{Metrics}
Performance is reported primarily using Average Accuracy \cite{tang2021graphbased_gcl}, which is defined as $Accuracy_{i, j}$ being the accuracy of Task j after training till Task i.

\begin{align}
    Average\ Accuracy = \frac{1}{T} \sum_{j=0}^{T} Accuracy_{T, j}
\end{align}

This is supplemented with Average Forgetting as defined by \cite{tang2021graphbased_gcl} as well.

\begin{align}
    Average\ Forgetting = \frac{1}{T-1} \sum_{j=0}^{T-1} (Accuracy_{T, j} -  Accuracy_{j, j})
\end{align}

Values reported for both metrics are obtained by averaging across 5 runs.

\subsection{Results on Online Class-Incremental Learning}
\begin{table}
    \centering
    \small
    \begin{tabular}{ccrr}
        \toprule
        Model Family  & Method & SVHN & CIFAR10 \\
        \midrule
        CNN & EWC  & 18.76$\pm$0.10          & 18.49$\pm$0.13 \\
            & GEM  & 33.40$\pm$0.27          & 22.88$\pm$4.06 \\
            & ER   & 45.51$\pm$3.27          & 29.94$\pm$2.08 \\
            & GCL  & \textbf{60.68$\pm$3.03} & 49.62$\pm$1.85 \\
        GNN & DGN  & 10.20$\pm$0.02          & 14.32$\pm$0.30 \\
            & Ours & 53.82$\pm$1.82          & \textbf{62.78$\pm$1.41} \\
        \bottomrule
    \end{tabular}
    \begin{tabular}{ccrr}
        \toprule
        Model Family  & Method & CIFAR100 & MiniImageNet \\
        \midrule
        CNN              & ER      & 20.50$\pm$0.90          & 11.00$\pm$0.50 \\
        (Pre-trained FE) & DER++   & 20.70$\pm$2.70          & 13.70$\pm$1.20 \\
                         & DualNet & 25.50$\pm$0.70          & 20.90$\pm$1.60 \\
                         & MuFAN   & 39.60$\pm$0.30          & 34.70$\pm$2.10 \\
        GNN              & DGN     &  8.35$\pm$0.64          &  7.86$\pm$0.53 \\
                         & Ours    & \textbf{43.87$\pm$1.02} & \textbf{41.06$\pm$1.92} \\
        \bottomrule
    \end{tabular}
    \caption{Comparison of average accuracy $\uparrow$ with CNN and GNN methods (left), and GNN and Pre-trained Feature Extractor methods (right)}
    \label{tab:main_acc}
\end{table}

The accuracy comparisons are shown in Table \ref{tab:main_acc}. Results for the CNN family of methods compared to SVHN and CIFAR10 are referenced from \cite{tang2021graphbased_gcl}. For CIFAR100 and MiniImageNet, we compare against the results from \cite{jung2023new_mufan} as they utilize pre-trained feature extractors as well. We adjust DGN \cite{Carta2021CatastrophicFI_dgn} for online Class-IL and report the best accuracies in Table \ref{tab:main_acc}. The full results with the various hyper-parameters for DGN are reported in Table \ref{tab:dgn}.

\begin{table}[t]
    \centering
    \begin{tabular}{clrrrr}
        \toprule
        Num Layer & Lambda & SVHN & CIFAR10 & CIFAR100 & MiniImageNet \\
        \midrule
        2 Layers & 0.01   & 10.16$\pm$0.11          & 13.75$\pm$0.75 &  7.09$\pm$0.75 &  7.30$\pm$1.13 \\
                 & 0.001  & 10.04$\pm$0.20          & \textbf{14.32$\pm$0.30} &  5.49$\pm$1.37 &  4.76$\pm$0.65 \\
                 & 0.0001 & 10.17$\pm$0.05          & 14.11$\pm$0.53 &  6.45$\pm$0.87 &  6.04$\pm$1.77 \\
        4 Layers & 0.01   & 10.05$\pm$0.16          & 14.00$\pm$0.33 & \textbf{ 8.35$\pm$0.64} &  7.41$\pm$1.84 \\
                 & 0.001  & \textbf{10.20$\pm$0.02} & 13.83$\pm$0.86 &  7.68$\pm$0.57 &  \textbf{7.86$\pm$0.53} \\
                 & 0.0001 & 10.10$\pm$0.16          & 14.04$\pm$0.27 &  7.87$\pm$0.62 &  7.43$\pm$0.93 \\
        \bottomrule
    \end{tabular}
    \caption{DGN Replication}
    \label{tab:dgn}
\end{table}

On CIFAR10, CIFAR100, and MiniImageNet, FGAT outperforms the relevant baseline methods by 26\%, 6\%, and 18\%, respectively. However, on the SVHN dataset, while FGAT achieves superior performance compared to methods like EWC \cite{doi:10.1073/pnas.1611835114_ewc}, GEM \cite{LopezPaz2017GradientEM_gem_splitcifar100}, and ER \cite{Chaudhry2019ContinualLW_er}, it falls short of matching the performance of GCL.

\begin{figure}[t]
    \centering
    \includegraphics[width=0.3\linewidth]{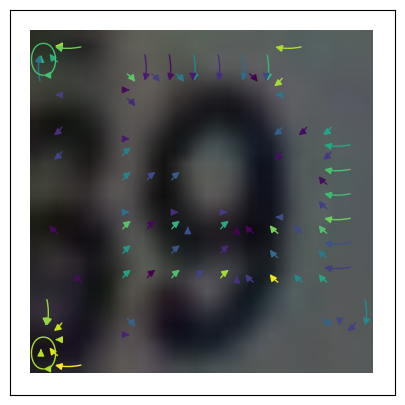}
    \includegraphics[width=0.65\linewidth]{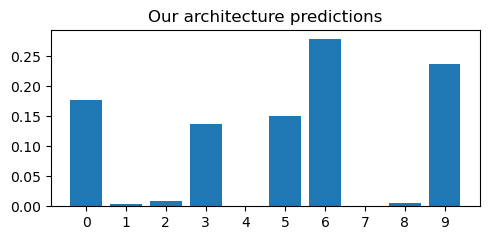}
    \caption{Visualization of an attention heads's top 95\% attention weights for SVHN image utilizing final trained model (left),  the prediction probability (right)}
    \label{fig:svhn_visualization}
\end{figure}

To understand why FGAT underperforms GCL on the SVHN dataset, we analyzed the edges with the top 95\% of attention scores for incorrectly predicted images. An example is provided in Figure \ref{fig:svhn_visualization}. Our analysis revealed that in SVHN, many images often contain more than one number, but are assigned a single number label due to the dataset construction method, which is inaccurate to the content of the image. As shown in Figure \ref{fig:svhn_visualization}, while the defined image class is 9 (the primary number centered in the image), a nearly complete 6 (a noisy, secondary number) is also visible. This labeling inconsistency likely confuses the model, as it assigns attention to both numbers, thereby impacting its ability to correctly classify the primary number. 

We also compare average forgetting on SVHN and CIFAR10 against the results reported by \cite{tang2021graphbased_gcl}, as shown in Table \ref{tab:main_fgt}. Similar to average accuracy, FGAT outperforms the compared methods on CIFAR10 significantly. However, while FGAT surpasses most of the compared CNN-based methods, it still falls short of GCL due to the data labeling issue in the original dataset.

\begin{table}[t]
    \centering
    \small
    \begin{tabular}{ccrr}
        \toprule
        Model Family  & Method & SVHN & CIFAR10 \\
        \midrule
        CNN & EWC &         94.99$\pm$1.23  & 86.95$\pm$1.15 \\
            & GEM &         68.91$\pm$4.06  & 76.90$\pm$5.53 \\
            & ER  &         62.37$\pm$4.33  & 72.64$\pm$4.88 \\
            & GCL & \textbf{21.86$\pm$2.35} & 35.69$\pm$3.33 \\
        GNN & DGN & 59.83$\pm$0.70 & 71.02$\pm$0.89 \\
        \begin{tabular}{@{}c@{}} \end{tabular} & Ours & 32.03$\pm$4.17 & \textbf{23.96$\pm$2.07} \\
        \bottomrule
    \end{tabular}
    \caption{Comparison of average forgetting $\downarrow$}
    \label{tab:main_fgt}
\end{table}

\subsection{Ablation Study} \label{subsec:ablation}
\begin{figure}[t]
    \centering
    \includegraphics[width=0.9\linewidth]{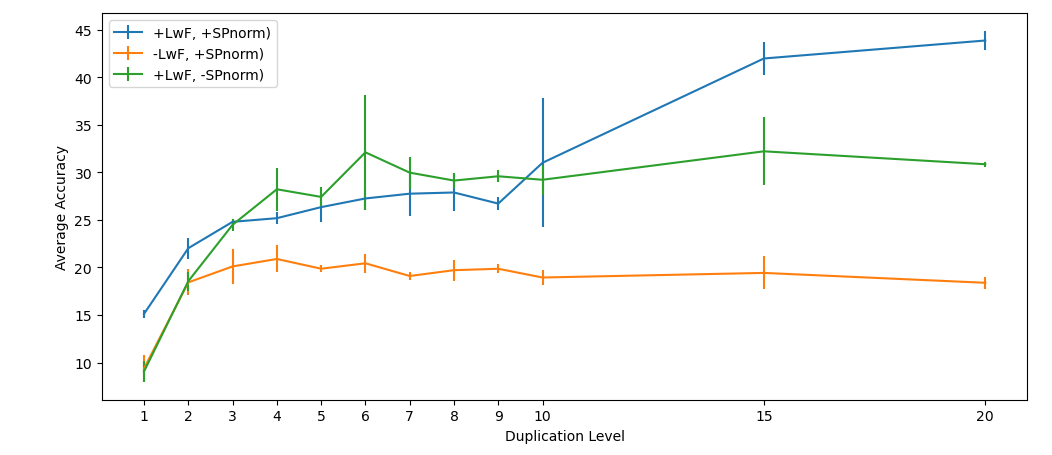}
    \caption{Effect of regularization, normalization \& duplication level}
    \label{fig:main_ablation_graph}
\end{figure}

\textbf{Duplication Level, Regularization and Normalization.} We further evaluate the pipeline's performance on CIFAR100, testing with both regularization and normalization, as well as with either component omitted, across various duplication levels. The results are summarized in Table \ref{tab:duplication_level}, and the average accuracy for the different setups across duplication levels is illustrated in Figure \ref{fig:main_ablation_graph}. We observe that increasing the duplication level results in an increase in average accuracy. However, above duplication level 15, the accuracy begins to drop if either LwF or SPN is removed. In contrast, when both techniques are present, there is a synergistic effect and the average accuracy obtained continues to increase.

\begin{table}[t]
    \centering
    \small
    \begin{tabular}{cccc}
        \toprule
        Duplication & SPN only & LwF only & LwF+SPN \\
        \midrule
        x1   &  9.37$\pm$1.39 &  9.05$\pm$1.11 & 15.11$\pm$0.38\\
        x2   & 18.43$\pm$1.37 & 18.55$\pm$0.98 & 22.02$\pm$1.10 \\
        x3   & 20.10$\pm$1.82 & 24.48$\pm$0.64 & 24.80$\pm$0.30\\
        x4   & 20.89$\pm$1.41 & 28.22$\pm$2.27 & 25.18$\pm$0.65 \\
        x5   & 19.86$\pm$0.36 & 27.41$\pm$1.03 & 26.34$\pm$1.54 \\
        x6   & 20.43$\pm$0.98 & 32.11$\pm$6.01 & 27.25$\pm$1.24\\
        x7   & 19.10$\pm$0.43 & 29.97$\pm$1.67 & 27.75$\pm$2.35\\
        x8   & 19.71$\pm$1.09 & 29.14$\pm$0.82 & 27.88$\pm$1.93\\
        x9   & 19.86$\pm$0.46 & 29.59$\pm$0.61 & 26.71$\pm$0.71\\
        x10  & 18.94$\pm$0.77 & 29.22$\pm$1.60 & 31.00$\pm$6.77\\
        x15  & 19.42$\pm$1.73 & 32.21$\pm$3.56 & 41.98$\pm$1.70\\
        x20  & 18.38$\pm$0.63 & 30.85$\pm$0.27 & \textbf{43.87$\pm$1.02} \\
        \bottomrule
    \end{tabular}
    \caption{Impact of LwF, SPN, and duplication level on CIFAR100, with 600 images per class, split into 500 train and 100 test. Base number of rehearsal images per class is 10, scaling accordingly and is 200 at maximum duplication level of x20.}
    \label{tab:duplication_level}
\end{table}

\begin{table}[t]
    \centering
    \small
    \begin{tabular}{crr}
        \toprule
        Graph Pooling & CIFAR100 \\
        \midrule
        GlobalMaxPool           & 28.45$\pm$6.07\\
        GlobalAddPool           & 19.93$\pm$1.98\\
        GlobalMeanPool          & 42.12$\pm$1.09\\
        Weighted GlobalMeanPool & \textbf{43.87$\pm$1.02} \\
        \bottomrule
    \end{tabular}
    \caption{Impact of graph pooling method}
    \label{tab:graph_aggr}
\end{table}

\textbf{Graph Pooling.} We also evaluate the impact of the proposed weighted global mean pooling, comparing it against other global pooling methods available in PyTorch Geometric \cite{fey2019fastgraphrepresentationlearning_pyg}. The results, shown in Table \ref{tab:graph_aggr}, indicate that mean pooling generally performs better, with the weighted version achieving the highest average accuracy.

\textbf{Layer Type Attention Head.} Additionally, we investigate the effect of the layer type and number of attention heads across both tessellated and non-tessellated image forms. The results are presented in Table \ref{tab:layer_type} and Table \ref{tab:attention_heads} respectively. From Table \ref{tab:layer_type}, we observe that for CIFAR100, the attention mechanism significantly improves performance. However, from Table \ref{tab:attention_heads} the optimal number of attention heads is crucial, with performance degrading when the number exceeds three. Moreover, the tessellated form, which is more flexible and parameter-efficient, tends to yield better results.
Extending this tessellated form testing to SVHN and CIFAR10 which uses the non-tessellated form, we observe a performance drop from 53.11$\pm$4.10 to 38.91$\pm$4.90 on SVHN, and from 62.78$\pm$1.41 to 57.10$\pm$1.87 on CIFAR10. However, it is worth noting that despite the decrease in performance on CIFAR10, our method still outperforms the compared methods.

\begin{table}[t]
    \centering
    \small
    \begin{tabular}{ccrr}
        \toprule
        Aggregation & Node Weights & CIFAR100 \\
        \midrule
        Mean      & Learnable              & 40.61$\pm$1.24 \\
        Summation & Learnable              & 40.86$\pm$0.69 \\
        Summation & Learnable\&Node Degree & 42.38$\pm$0.36 \\
        Summation & Learnable\&Attention   & \textbf{43.87$\pm$1.02} \\
        \bottomrule
    \end{tabular}
    \caption{Impact of graph layer aggregation and type of node weights, layers are in order of: SageConv, GraphConv, GCNConv, GATv2Conv}
    \label{tab:layer_type}
\end{table}

\begin{table}[t]
    \centering
    \small
    \begin{tabular}{crr}
        \toprule
         Num Attention Heads & Non-tessellated  & Tessellated \\
        \midrule
        1  & 38.33$\pm$3.85 & 43.16$\pm$1.31 \\
        2 & 34.45$\pm$5.25 & 43.54$\pm$0.71 \\
        3  & 28.66$\pm$1.61 & \textbf{43.87$\pm$1.02} \\
        4  & 27.61$\pm$0.44 & 38.70$\pm$4.16 \\
        5  & 30.63$\pm$6.00 & 34.66$\pm$4.97 \\
        6  & 27.50$\pm$0.80 & 34.05$\pm$4.46 \\
        7  & 28.03$\pm$1.76 & 30.35$\pm$4.89 \\
        8  & 35.04$\pm$6.00 & 32.52$\pm$5.77 \\
        \bottomrule
    \end{tabular}
    \caption{Impact of number of attention heads}
    \label{tab:attention_heads}
\end{table}

\textbf{Model Size Comparison}
A summary of the model parameter counts is presented in Table \ref{tab:model_size}. The model sizes are compared against the traditional ResNet family of models \cite{He2015DeepRL_resnet}, the Vision Transformer (ViT) family of models \cite{dosovitskiy2020vit}, as well as the model sizes reported by \cite{jung2023new_mufan}. Reference parameter counts were obtained from \cite{Ansel_PyTorch_2_Faster_2024_pytorch}.

We observe a notable increase in the number of parameters when utilizing the non-tessellated form of the weighted global mean pooling. This increase is attributed primarily to the weighing layer parameters, which grow from 2,401 to 38,416. Despite this, even the largest configuration of the model, while exceeding the size of ResNet18, is less than half the size of ViT-Base. If the tessellated form is used instead—offering competitive performance on CIFAR10, CIFAR100, and MiniImageNet—the number of parameters remains lower than MuFAN \cite{jung2023new_mufan} and only slightly higher than ResNet18.

In practice, a significant portion of these parameters are not trainable, as the feature extractor is frozen. Excluding the approximately 11 million parameters of the feature extractor, the number of trainable parameters amounts to roughly 1\% of the smallest ViT model, demonstrating the model's efficiency in parameter utilization.

\begin{table}[t]
    \centering
    \begin{tabular}{crr}
        \toprule
        Family & Model & Parameters \\
        \midrule
        ResNet & 18      & 11M   \\
               & 34      & 21M   \\
               & 50      & 25M   \\
               & 101     & 44M   \\
               & 152     & 60M   \\
        Ours  & 3A-GMeP  & 12M  \\
              & 3A-WTes  & 13M  \\
              & 3A-NTes  & 26M \\
              & 4A-GMeP  & 12M  \\
              & 4A-WTes  & 13M  \\
              & 4A-NTes  & 32M \\
        MuFAN & -        & 14M \\
        ViT   & Base 16  & 86M   \\
              & Base 32  & 88M \\
              & Large 16 & 304M  \\
              & Large 32 & 306M \\
              & Huge 14  & 633M  \\
        \bottomrule
    \end{tabular}
    \caption{Model parameter count in millions, count for ours includes count for frozen feature extractor (11M)}
    \label{tab:model_size}
\end{table}

\section{Conclusion}
In this paper, we proposed a novel framework for online continual learning in image classification, leveraging Graph Attention Networks (GATs) to capture contextual relationships and dynamically update task-specific representations. By transforming images into graph representations through hierarchical feature maps extracted by a pre-trained feature extractor, our method effectively encodes multi-granular semantic and spatial information. The incorporation of an enhanced global pooling strategy further improves classification performance, while the rehearsal memory duplication technique addresses representation imbalances between current and past tasks without increasing the memory budget. Comprehensive experiments on four benchmark datasets, demonstrate the superiority of our approach over existing state-of-the-art methods. The results highlight the effectiveness of FGAT in continual learning scenarios, offering a robust and resource-efficient alternative to traditional CNNs and transformer-based models. 

\clearpage

\printbibliography

\clearpage

\end{document}